\documentclass{article}
\usepackage{amsmath,epsfig}
\usepackage[preprint]{spconfa4}
\usepackage{graphicx}
\usepackage{amsmath}
\usepackage{amssymb}
\usepackage{hyperref}       % hyperlinks
\usepackage{cleveref}
\usepackage{flushend}

\copyrightnotice{978-1-6654-3864-3/21/\$31.00 ©2021 IEEE}

\let\OLDthebibliography\thebibliography
\renewcommand\thebibliography[1]{
  \OLDthebibliography{#1}
  \setlength{\parskip}{0pt}
  \setlength{\itemsep}{0pt plus 0.3ex}
}

\begin{document}\sloppy

% Example definitions.
% --------------------
\def\x{{\mathbf x}}
\def\L{{\cal L}}

% Title.
% ------
\title{Weakly-supervised Audio-visual Sound Source Detection and Separation}
%
% Address.
% ---------------
\name{Tanzila Rahman$^{1,2}$ and Leonid Sigal$^{1,2,3}$}
\address{$^1$University of British Columbia \qquad
$^2$Vector Institute for AI \qquad
$^3$Canada CIFAR AI Chair \\
{\tt\small \{trahman8, lsigal\}@cs.ubc.ca}
}

\maketitle

\begin{abstract}
Learning how to localize and separate individual object sounds in the audio channel of the video is a difficult task. Current state-of-the-art methods predict audio masks from artificially mixed spectrograms, known as Mix-and-Separate framework. We propose an audio-visual co-segmentation, where the network learns both what individual objects look and sound like, from videos labeled with only object labels. Unlike other recent visually-guided audio source separation frameworks, our architecture can be learned in an end-to-end manner and requires no additional supervision or bounding box proposals. Specifically, we introduce weakly-supervised object segmentation in the context of sound separation. We also formulate spectrogram mask prediction using a set of learned mask bases, which combine using coefficients conditioned on the output of object segmentation — a design that facilitates separation. Extensive experiments on the MUSIC dataset show that our proposed approach outperforms state-of-the-art methods on visually guided sound source separation and sound denoising.
\end{abstract}
\vspace{-0.10in}
\section{Introduction}
\label{sec:intro}
\vspace{-0.10in}
Multi-modal, visual and auditory, perception is an important research topic. Human brain has remarkable ability to isolate specific conversation from a noisy environment, as noted by Cherry through ``cocktail party effect"~\cite{cherry1953some}. At the same time, we can recognize objects and segment regions corresponding to those objects using our visual and auditory systems. We can also imagine how a particular, visually depicted, object may sound. Each object has unique physical properties, some of which can be visually observed, which leads it to generate a unique sound modulated by interactions with other objects and the environment. Therefore, working jointly with auditory and visual cues can be very useful for recognition of objects, localization of object regions and separation of sounds they make. % in an unsupervised way. 
Separating sounds of each object from a video has wide range of applications including audio denoising, hearing aids, automated transcription of speech and music, equalization, audio event remixing and dialog following. %  and so on~\cite{gao2019co}.

\begin{figure}[t]
  \centering
  \includegraphics[scale=0.31]{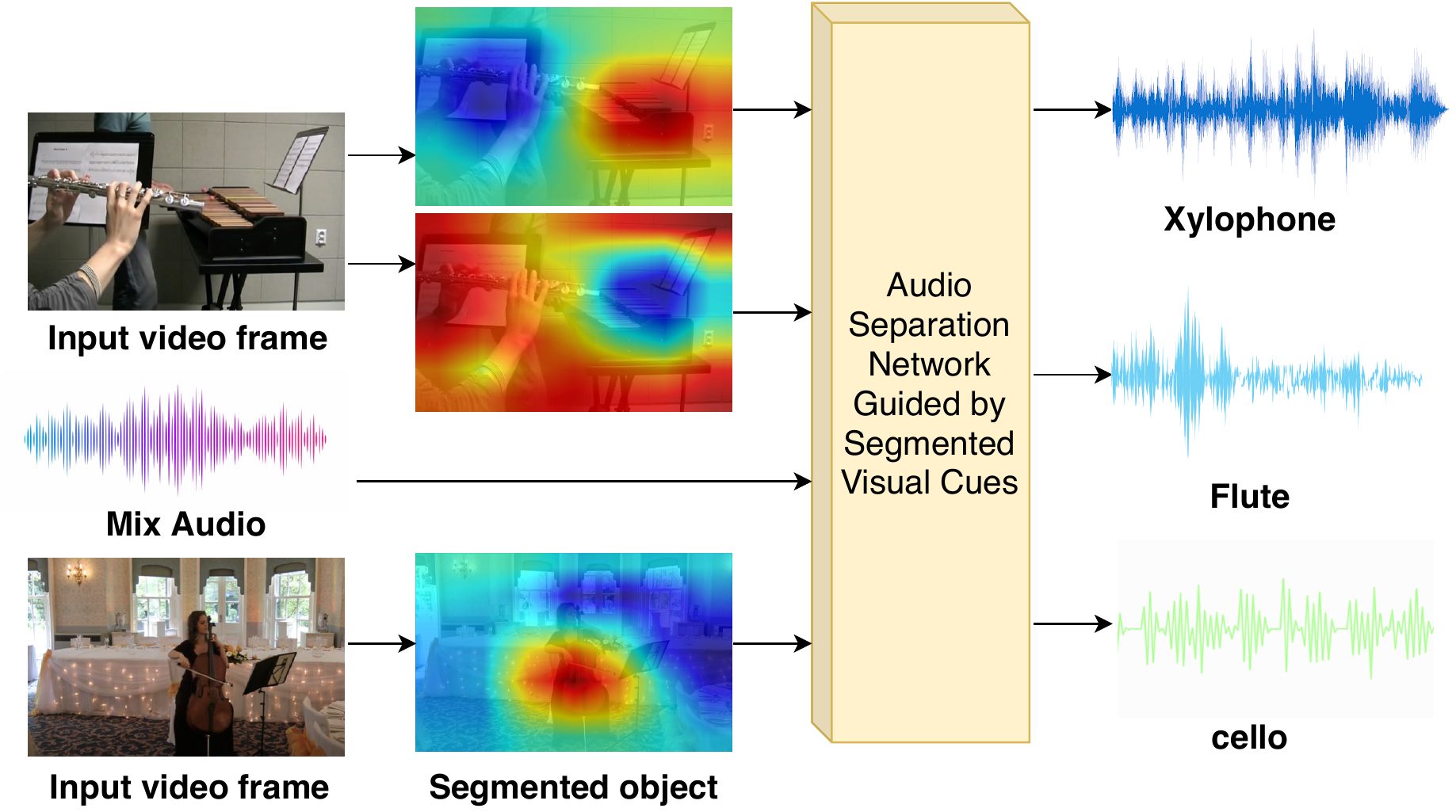}
  \vspace{-0.10in}
  \caption{{\bf Visually-guided Sound Source Separation.} Our method first detects and segments object(s) which make sound(s), in a weakly-supervised manner, and then separates their respective audio signals.}%  sound of each individual object.}
  \label{fig:intro}
  \vspace{-0.3in}
\end{figure}

Recent methods for audio-visual source separation~\cite{gao2019co, zhao2019sound, zhao2018sound} utilize “mix-and-separate” approach to train neural network architectures using self-supervision. The paradigm is simple, given a video, mix the audio track by combining audio channel with one from another video, and train the network to recover the original audio back, conditioned on the visual encoding of corresponding video content. This paradigm effectively synthesizes “cocktail party effect” by mixing clean sound(s) with others not present in the scene. While effective in training models for variety of tasks, such as sound source separation~\cite{gao2019co, zhao2018sound} and on-/off-screen audio identification~\cite{owens2018audio}, this approach implicitly assumes that videos contain single-source sounds and attempts to correlate regions of the video with spectrograms \cite{zhao2018sound}. Co-separation approach recently introduced by Gao {\em et al.} \cite{gao2019co} addresses single-source limitation, but relies on object detectors trained with an external dataset %(Open Images \cite{openimages}) 
annotated with bounding  boxes for potential audible objects. %However, object annotations are generally difficult and expensive to collect. 
In addition, while audio classes and corresponding spectrogram segmentations, that correspond to detected regions, are “discovered” during training, the model has no capacity to refine object detectors themselves to be optimal for sound-source separation task; {\em e.g.}, an entire object region is implicitly assumed to produce the sound. 

Inspired by prior work,  we aim to address aforementioned limitations. Specifically, we propose a weakly-supervised audio-visual detection and separation method. Our approach, similar to \cite{gao2019co}, does not assume single-source video; but, unlike \cite{gao2019co}, also does not rely on externally trained object detection module or object-level annotations of any kind. Instead, we leverage weak video-level labels to jointly learn visual and auditory segmentors that depend on one another. Our architecture has two paths: (1) a {\em video frame semantic segmentation path} designed to segment a frame into a set of regions using an attention mechanism that generates per-object-class attention map trained using weak frame-level classification objective; and (2) a {\em spectrogram mask prediction path} which takes both mixed spectrogram and pooled object-class image features and outputs a dense spectrogram mask with an objective to mask out the mixed-in sound. The spectrogram mask prediction branch is implemented using attention U-Net architecture \cite{oktay2018attention}, similar to \cite{gao2019co, zhao2018sound}. However, importantly, unlike prior methods, we train U-Net to produce a set of base masks from which a final mask is constructed using a set of sparse coefficients predicted from multi-modal audio-visual features. This architecture design takes inspiration from \cite{bolya2019yolact}. We find that such bi-linear decomposition is very useful in practice, allowing spectrograms to collaborate in learning a set of auditory sound bases, while relying on coefficient predictor to figure out how those should mix for a specific object type. Finally, despite having weaker supervision (no object annotations), compared to \cite{gao2019co}, we illustrate superior performance on the benchmark MUSIC and cross-dataset performance on AudioSet datasets.

\vspace{0.03in}
\noindent
{\bf Contributions.} Our main contribution is an audio-visual co-segmentation approach for sound source separation, where the network learns both what individual objects look like and sound like, from videos labeled with only, one or more, object labels. This formulation and architecture has a number of appealing properties. Mainly, it does not assume single sound source input data, can be learned in an end-to-end manner, and requires no additional supervision or bounding box proposals. On the technical side, we introduce weakly-supervised object segmentation in the context of sound separation. We also formulate spectrogram mask prediction using a set of learned mask bases which are combined using sparse coefficients  conditioned on multi-modal (visual object and auditory) features. Extensive experiments on the MUSIC dataset \cite{zhao2018sound} show that our proposed approach outperforms state-of-the-art methods on visually guided sound source separation and sound denoising.
%------------------------------------------------------------------------
\vspace{-0.10in}
\section{Related Work}
\vspace{-0.10in}
\noindent
\textbf{Audio-only sound source separation.} 
Sound source separation is a challenging problem in speech processing and was first illustrated by ``cocktail party effect"~\cite{cherry1953some}. %Sound separation has a wide range of applications, including music/vocal separation~\cite{jansson2017singing, rafii2012repeating}, speech separation and enhancement~\cite{gabbay2018seeing, khonglah2016speech, li2019multichannel}. 
Classical approaches for the task include 
%Independent Component Analysis (ICA)~\cite{hyvarinen2000independent}, sparse decomposition~\cite{zibulevsky2000blind}, Computational Auditory Scene Analysis (CASA)~\cite{ellis1996prediction}, probabilistic latent variable models~\cite{hofmann1999probabilistic, shashanka2008probabilistic}, 
local Gaussian modeling~\cite{fitzgerald2016projet}, %kernel additive modeling~\cite{liutkus2014kernel}, 
and Non-negative Matrix Factorization (NMF)~\cite{le2015deep}.
%Among traditional approaches, NMF is still widely used for unsupervised source separation~\cite{guo2015nmf, innami2012nmf, jaiswal2011clustering, spiertz2009source, virtanen2003sound}, but requires extra supervision to get good results. 
Recently, deep learning based approaches~\cite{gao2019co, zhao2019sound, zhao2018sound} have gained popularity and most of recent methods
%shown significant improvement in separation performance over earlier methods.  Most of recent methods~\cite{hershey2016deep, huang2015joint} 
use "Mix-and-Separate" framework to train the network by artificially mixing multiple audio streams first and then  learning to separate each audio from the mixture. We also use mix-and-separate idea, but %following~\cite{gao2019co, zhao2019sound, zhao2018sound}
use visual features to guide audio separation.

\vspace{0.03in}
\noindent
\textbf{Audio-visual source separation.} 
Multi-modal learning has recently become a popular topic in the computer vision community.
Auditory signal is used to supervise vision model during training in~\cite{owens2016ambient}.  Similarly, in~\cite{aytar2016soundnet} visual features are used to guide sound models. %Audio and vision jointly are used to train the model in~\cite{arandjelovic2017look, korbar2018co}. The same line of research also includes generating sounds from silent video~\cite{owens2016visually, zhou2018visual}.  
Following~\cite{zhao2018sound, xu2019recursive}, we use audio-visual features to perform the separation. Unlike~\cite{gao2019co}, we do not use any pre-trained object detector and propose an end-to-end approach to detect, localize and separate sound sources. %Early audio-visual works also explore the relation between motion and sound. Fisher {\em et al.}~\cite{fisher2001learning} used maximal mutual information, while others focused on canonical correlation methods \cite{izadinia2012multimodal, kidron2005pixels}.  Unlike~\cite{barzelay2007harmony, hu2018squeeze, zhao2019sound} who focus on motion and onset events, our aim is to detect and segment sound making objects to supervise sound separation task.

\begin{figure*}[!t]
  \centering
  \includegraphics[scale=0.41]{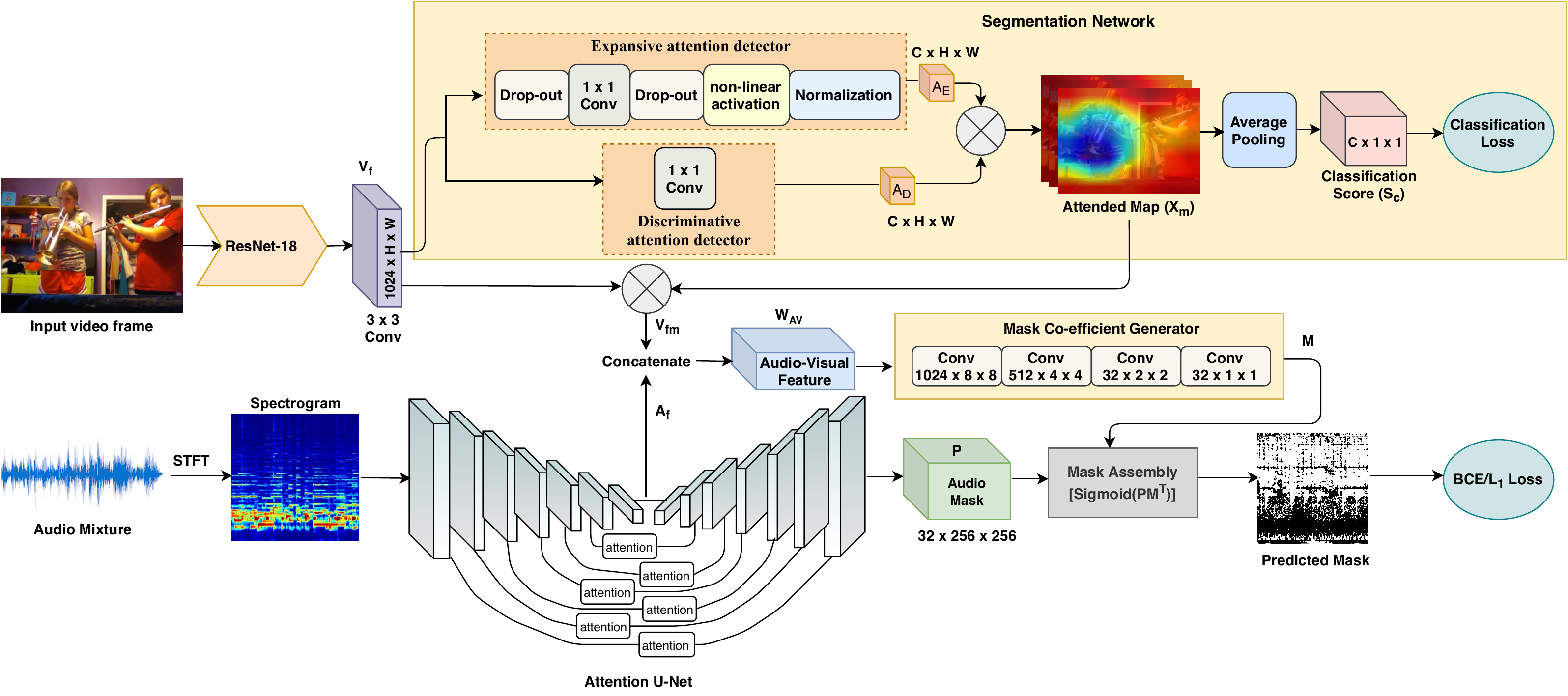}
  \vspace{-0.12in}  
  \caption{{\bf Weakly-supervised Audio-Visual Architecture.} ResNet-18 followed by a $3\times3$ convolution layer is used to extract visual feature ($\mathbf{V}_f$) from input video frames and fed to the segmentation network to detect the sound sources. Depending on the classification scores from the segmentation network, we generate soft semantic segmentations by producing class-specific attention map ($\mathbf{X}_m$). We use this attention map to pool features from respective image regions of $\mathbf{V}_{f}$ generating $\mathbf{V}_{fm}$. The resultant feature is concatenated with the bottleneck features of attention u-net to generate audio-visual feature ($\mathbf{W}_{AV}$). $\mathbf{W}_{AV}$ is passed to the mask coefficient generator to generate $k$ mask coefficients ($\mathbf{M}$). At the same time, attention U-Net generates $k$ audio channels ($\mathbf{P}$) and combined linearly ($\sigma(\mathbf{P}\mathbf{M}^T)$) to predict final audio spectrogram mask guided by visual feature. }
  \label{fig:overview}
  \vspace{-0.17in}
\end{figure*}

\vspace{0.03in}
\noindent
\textbf{Weakly supervised visual learning.}
Given a video, our approach is able to detect which audio signals correspond to which objects and localize those objects within the video frames in a weakly-supervised manner.
% , which requires no proposals or any instance-level supervision. %Instead, only video-level labels are required. 
Earlier approaches address weakly-supervised object detection and segmentation using Multiple Instance Learning (MIL)~\cite{pinheiro2015image}. 
% and Expectation-Maximization (EM) procedure~\cite{papandreou2015weakly}. 
More recently, pseudo-annotation generation~\cite{shen2017weakly} has gained popularity. %Feedback CNN architectures can use CAM \cite{zhou2016learning} or Grad-CAM \cite{selvaraju2017grad} to identify discriminative object regions directly, or as a form of pseudo-annotations. % for strong fully-supervised methods like Faster R-CNN \cite{ren2015faster} or Mask R-CNN \cite{he2017mask}, leading to a two-stage approach. Zhang {\em et al.}~\cite{zhang2019decoupled} propose an end-to-end pseudo-annotation generation pipeline by introducing a decoupled spatial neural attention network to localize discriminative parts and estimate object regions simultaneously. 
Motivated by~\cite{zhang2019decoupled}, we generate pseudo-annotations for weakly-supervised segmentation; and use these to visually guide the network to separate sound in an end-to-end fashion instead of using pre-trained object detectors~\cite{gao2019co}.

%-------------------------------------------------------------------------

%
\vspace{-0.10in}
\section{Approach}
\vspace{-0.10in}
We introduce a method for visually-guided sound separation, which leverages segmented object regions predicted to make sounds. In this section we first formalize our audio-visual sound source separation and detection task (Section~\ref{problem_formulation}) and then focus on describing the proposed deep neural network architecture for solving it (Section~\ref{model_architecture}).

\vspace{-0.10in}
\subsection{Problem Formulation}
\label{problem_formulation}
\vspace{-0.10in}
%Given an unlabeled video $V$ with accompanying audio $A(t)$ containing a set of $N$ objects denoted as $V={O_1, O_2, ...., O_N}$. Objects within the video can be treated as sound sources and $A(t) = \sum_{n=1}^N A_n(t)$ where $A_n(t)$ is the discrete time signal for each object. In this work, our goal is to detect and separate sound $A_n(t)$ of each object $O_N$ by using object level audio-visual supervision.

In this work, we use "Mix-and-Separate" framework~\cite{gao20192, gao2019co, zhao2018sound}, a well known approach for the task of sound source separation. The idea is to generate an artificially complex auditory signal by mixing multiple individual audio signals and learn to separate each individual sound of interest from the composition (see Figure~\ref{fig:intro} for illustration). 
% We follow a similar strategy for our training.  

Given two input videos $V_1$ and $V_2$ with accompanying audio $A_1(t)$ and $A_2(t)$, respectively,  we detect and segment objects, that make sound, from each video using a weakly supervised segmentation network. Then we generate a complex mixed auditory signal $A_m(t) = A_1(t) + A_2(t)$ by mixing two audio signals $A_1(t)$ and $A_2(t)$. Using a short-time Fourier transform (STFT)~\cite{griffin1984signal} with $F$-frequency bins, we transformed the mixed signal $A_m(t)$ into a magnitude spectrogram $A^M \in \mathbb{R}_+^{F \times N}$. $A^M$ represents the change of frequency and phase over the time in mixed auditory signal. Suppose $V_1$ contains two objects $O_1'$ and $O_1''$ and corresponding audios $A_1(t)'$ and $A_1(t)''$ accordingly. Similarly, $V_2$ contains one object $O_2$ with accompanying audio $A_2(t)$. Now our goal is to separate sounds $A_1(t)'$, $A_1(t)''$ and  $A_2(t)$ of each detected object $O_1'$, $O_1''$ and $O_2$ by predicting a spectrogram mask $\mu_n$ with the supervision of visual cues. To train the network one can use either ratio or binary mask and obtain object level magnitude spectrogram by $A_n = A^M \times \mu_n$. Finally, one can apply Inverse Short-Time Fourier transform (ISTFT)~\cite{griffin1984signal} to reconstruct object level wave-form sounds.

\vspace{-0.15in}
\subsection{Weakly-supervised Audio-Visual Architecture}
\label{model_architecture}
\vspace{-0.10in}
We propose a weakly-supervised audio-visual detection and separation architecture illustrated in Figure~\ref{fig:overview}. Our architecture has two paths: (1) a {\em video frame semantic segmentation path} designed to detect objects that have potential to make sounds and segment them out in the frame, using an attention mechanism that generates per-object-class attention map, trained using weak frame-level classification objective (top block in yellow in Figure~\ref{fig:overview}); and (2) a {\em spectrogram mask prediction path} which takes both mixed audio and pooled object-class image features and outputs a dense mask with an objective to mask out the mixed-in sound (bottom block, Figure~\ref{fig:overview}). 

We propose an end-to-end approach, unlike~\cite{gao2019co}, to detect and segment objects from the input video frame. % Implicitly, we treat the detected and segmented object as a source of sound. 
The input to our video frame segmenter is an RGB image/frame. The output is two fold -- (i) a one-channel semantic segmentation attention map, per object class, that highlights regions where this object is present and (ii) probability of this object being present in the first place. Note, that (i) is only meaningful for objects that are present (probability of presence is  high, above a  threshold $\tau$). 

The spectrogram mask prediction path is trained to generate a (binary or real-valued) mask that masks-out the mixed-in sound. Prior approaches decode the multi-modal encoding of the mixed-audio and visual representation of attended frame \cite{zhao2018sound}, or an object region in the frame \cite{gao2019co}, into a mask directly. Instead, we utilize an attention U-Net architecture to first dynamically generate auditory mask bases from the mixed spectrogram itself. We then generate coefficients for these bases conditioned on the multi-modal features. The final mask is constructed as a coefficient-weighted combination of predicted bases. This decomposition allows shared learning of bases, and focuses visual conditioning on a few coefficients; this, we find, significantly improves the performance. 

\vspace{0.03in}
\noindent
\textbf{Video frame semantic segmentation.}
We use ResNet-18~\cite{he2016deep} as backbone network followed by a $3\times3$ \textit{convolution} to extract $H \times W$ spatial visual features $\mathbf{V}_f \in \mathbb{R}^{1024 \times H \times W}$ from the input video frame. These features are feed to the segmentation network to detect and segment objects. Following~\cite{zhang2019decoupled}, our object detection network uses a decoupled spatial neural attention to detect and localize salient object regions simultaneously. 
% As a results, network is able to generate multi-class semantic segmentation pseudo-annotations in a weakly supervised manner. 
%
The segmentation network contains two branches: (1) Expansive attention detector which identifies object regions and generates expansive attention map $\mathbf{A}_E \in \mathbb{R}^{C\times H\times W}$; and (2) Discriminative attention detector that predicts the discriminative parts and generates discriminative attention map $\mathbf{A}_D \in \mathbb{R}^{C\times H\times W}$. Expansive attention detector consists of a drop-out layer, $1\times1$ \textit{convolution layer}, another drop-out layer, a non-linear activation layer (Eq.~\ref{eq:non_linear_activation}) and  a spatial-normalization step (Eq.~\ref{eq:spatial_normalization}).
Each element in $\mathbf{A}_E$ is defined as follows: 
\vspace{-0.03in}
\begin{equation} \label{eq:non_linear_activation}
    \alpha_{(i,j)}^c = F(\mathbf{W}_c^T\mathbf{V}_{f}(:, i,j) + b^c),
\end{equation}
\begin{equation} \label{eq:spatial_normalization}
    \alpha_{(i,j)}^c = \frac{\alpha_{(i,j)}^c}{\sum_{i}^{H}\sum_{j}^{W} \alpha_{(i,j)}^c},
\end{equation}
where $c \in \mathcal{C}$ and $F(\cdot)$ denote  channel/class and non-linear activation respectively. Discriminative attention detector contains a $1\times1$ \textit{convolution layer} and directly outputs a class-specific object attention map $\mathbf{A}_D$. 
%  generates attended feature $\beta_{(i,j)}^c$.  
We combine both attentions and generate final attention maps as follows: $\mathbf{X}_m = \mathbf{A}_E \odot \mathbf{A}_D$, where $\odot$ is the element-wise multiplication. 
%
% \begin{equation}
%    \mathbf{X}_m = \alpha_{(i,j)}^c \beta_{(i,j)}^c
%\end{equation}
%
Each depth channel of $\mathbf{X}_m$ is passed through a \textit{spatial average pooling layer} to generate classification score for corresponding class; this results in $\mathbf{S} \in \mathbb{R}^{|\mathcal{C}|}$ class scores. Then we apply a multi-label classification loss (\textit{c-loss}) denoted as follows:
\vspace{-0.05in}
\begin{small}
\begin{equation} 
    \mathcal{L}_{c-loss} = -\sum_c^{\mathcal{C}} \left[ \mathbf{y}_c \log\frac{1}{1+e^{-\mathbf{S}_c}} + (1-\mathbf{y}_c)\log\frac{e^{-\mathbf{S}_c}}{1+e^{-\mathbf{S}_c}} \right],
\end{equation}
\end{small}
where $\mathbf{y}_c$ denotes binary GT label for corresponding $c$-th class and $|\mathcal{C}|$ is the number of object classes. 

Note that $\mathbf{X}_m$ can be interpreted as soft semantic segmentation (segmentation can be obtained by thresholding $\mathbf{X}_m^{c}$), with each channel corresponding to a specific object type. We can detect which objects are present, at test time, in a given video frame, by thresholding the classification scores $\mathbf{S}$. 

\vspace{0.04in}
\noindent
\textbf{Attention U-Net for audio processing.}
Motivated by~\cite{zhao2018sound}, in this work, we use time-frequency representation of sound. Therefore, first we apply STFT on the input mixture sound to generate corresponding spectrogram. Then magnitude of spectrogram is transformed into log-frequency scale and used for further processing. Following~\cite{oktay2018attention}, we use attention U-Net to extract audio features from the log magnitude of spectrogram. Attention U-Net uses attention gate (AG) % (see Figure~\ref{fig:attention_gate}) 
to highlight discriminative features while passing through the skip connection. 
We use 7 convolutions (or down-convolutions) and 7 de-convolutions (or up-convolution) with skip connections in between for attention U-Net. The size of input spectrogram is $1\times 256 \times 256$ and the final output of attention U-Net are audio mask bases ($\mathbf{P} \in \mathbb{R}^{k \times 256 \times 256}$) with $k$ channels/bases. In this work, we use $32$ as the value of $k$.

\begin{table*}[h]
\small
   \begin{center}
   \caption{{\bf Audio separation results on MUSIC test set.} Performance reported using SDR/SIR/SAR. SDR and SIR capture separation accuracy; SAR only captures the absence of artifacts.}
   \vspace{0.03in}
   \label{table:ex_results_final}
  \begin{tabular}{l|ccc|ccc} \hline
  & \multicolumn{3}{c|}{Ratio Mask} & \multicolumn{3}{c}{Binary Mask}\\
  \hline
     Methods & SDR ($\uparrow$) & SIR ($\uparrow$) & SAR ($\uparrow$) & SDR  ($\uparrow$) & SIR ($\uparrow$) & SAR ($\uparrow$) \\
    \hline 
    NMF-MFCC~\cite{spiertz2009source} & 0.92 & 5.68 & 6.84 & - & - & - \\
    AV-Mix-and-Separate~\cite{gao2019co} & 3.23 & 7.01 & 9.14 & - & - & - \\
    Sound-of-Pixels~\cite{zhao2018sound} & 7.81 & 11.06 & 14.05 & 7.26 & 12.25 & 11.11 \\
    CO-SEPARATION~\cite{gao2019co} & 7.64 & 13.8 & 11.3 & - & - & - \\
    \hline
    Ours (Mask coefficient) & 8.40 & 12.53 & 14.11 & 9.25 & \textbf{15.98} & \textbf{12.45} \\
    Ours (Mask coefficient + Seg. Net) & \textbf{9.14} & \textbf{13.35} & \textbf{14.18} & \textbf{9.29} & 15.09 & 12.43\\
    \hline
  \end{tabular}
  \end{center}
  \vspace{-0.12in}
  \vspace{-0.13in}
\end{table*}

\vspace{0.03in}
\noindent
\textbf{Mask Coefficient Generator.}
Following~\cite{bolya2019yolact}, the goal of mask coefficient generator is to predict $k$ mask coefficients: $\mathbf{M} \in \mathbb{R}^k$. In this work, we use audio-visual feature to generate mask coefficient. Based on classification scores, $\mathbf{S}_c$, that are  above a certain threshold, $\tau$, from the segmentation network, we select corresponding class-specific attention channel(s) of $\mathbf{X}_m$ and apply weighted pooling on the visual feature $\mathbf{V}_f$ to generate attended visual feature for a corresponding object -- $\mathbf{V}_{fm}$. The attended visual feature is concatenated with bottleneck U-Net feature, $\mathbf{A}_f$, to produce audio-visual feature vector $\mathbf{W}_{AV}$. $\mathbf{W}_{AV}$ is fed to the mask coefficient generator to predict $k$ mask coefficient ($\mathbf{M}$). The mask coefficient generator consists of a series of \textit{convolution layers} with non-linear activations and batch-normalization. In this paper, we use \textit{ReLU} as non-linear activation function. We predict final magnitude of spectrogram, $\mu_A$, by linearly combining $k$ audio mask bases from $\mathbf{P}$ with the mask coefficient $\mathbf{M}$:
%  as follows:
%
\begin{equation}
    \mu_A = \sigma(\mathbf{P} \mathbf{M}^T).
\end{equation}
The predicted magnitude of spectrogram $\mu_A$ is combined with the phase of input spectrogram. Then we use the inverse STFT to get a wave-form of the prediction. Our ultimate goal is to learn spectrogram masks of two types: {\em binary} or {\em ratio}. Following~\cite{zhao2018sound}, in case of binary mask we use per-pixel sigmoid cross entropy loss ({\em i.e.}, BCE Loss, $\mathcal{L}_{BCE}$, to train the network). Similarly, per-pixel $\mathcal{L}_1$ loss~\cite{zhao2016loss} is used to train the network when we use ratio mask.

\vspace{-0.10in}
\section{Experiments}
\vspace{-0.10in}

\subsection{Datasets}
\vspace{-0.05in}
\noindent
\textbf{MUSIC dataset.}
We evaluate our method using MUSIC %(Multimodal Sources of Instrument Combinations) 
dataset~\cite{zhao2018sound} which contains 685 untrimmed videos of musical solos and duets. %which includes 11 instrument categories: accordion, acoustic guitar, cello, clarinet, erhu, flute, saxophone, trumpet, tuba, violin and xylophone. MUSIC dataset provides YouTube videos by keyword query. We use these keywords to download videos. 
We find that $31$ videos are now missing from YouTube. The train/val/test split of MUSIC dataset is unavailable. Therefore, we follow the train/val/test split of~\cite{gao2019co} where the first/second video in each category is considered as the validation/test data, and the rest used for training data.

\vspace{0.03in}
\noindent
\textbf{AudioSet-SingleSource.} 
This is a small dataset, assembled in~\cite{gao2018learning}, which we only use for evaluation. %Dataset contains single object sounds with corresponding videos. 
The dataset consists of $15$ musical instruments plus additional sounds produced by animals and vehicles. For our cross-dataset experiment we randomly select $11$ out of $15$ musical instruments for evaluation. Note, number of the instruments are {\em unseen} by the model -- not in the MUSIC dataset that we use for training. 
\vspace{-0.03in}

\begin{table}[!h]
\small
  \begin{center}
  \vspace{-0.15in}
  \caption{{\bf Multi-label object classification accuracy.} Performance in (\%) on the MUSIC test set.}
  \label{table:classification_results}
  \vspace{0.03in}
  \begin{tabular}{l|ccccc} \hline
    Threshold value($\tau$) & 0.1 & 0.2 & 0.3 & 0.4 & 0.5\\
    \hline
    Binary mask & 80.30 & 91.41 & 93.18 & 92.68 & 88.89 \\
    Ratio mask & \textbf{83.08} & \textbf{91.92} & \textbf{93.69} & \textbf{93.18} & \textbf{89.65} \\
    \hline
  \end{tabular}
  \end{center}
  \vspace{-0.15in}
  \vspace{-0.05in}
\end{table}

\vspace{-0.15in}
\subsection{Pre-processing and implementation details}
\vspace{-0.05in}
Following~\cite{zhao2018sound}, %we use several pre-processing steps on the MUSIC dataset before training the model. 
to reduce the computational cost, we sub-sampled the audio signals to 11kHz %, so that most of the important frequencies of instruments will preserve by degrading slightly the overall audio quality. We 
and sample approximately 6 secs audio by random cropping from each untrimmed video. A Hann window size of 1022 and a hop length of 256 is used to compute STFT and generate a $512\times256$ Time-Frequency audio spectrogram which is further re-sampled on a log-frequency scale to obtain a $256\times256$ Time-Frequency representation. This representation is used as input to the attention U-Net. % module for further processing. 
We obtain an output predicted mask and apply an inverse sampling step to convert the mask back to linear frequency scale of size $512\times256$ followed by an inverse STFT to recover wave-form signal. %We use Pytorch to implement our network. 
Following~\cite{gao2019co}, we randomly sample 1-frame to train the model. To process the input video frame, we use ResNet-18~\cite{he2016deep} pre-trained on ImageNet. %with two modifications: (1) remove last average pooling layer and {\tt fc} layer; and (2) add a $3\times3$ convolution layer with 1024 output channels. 
We follow the experimental protocol of~\cite{zhao2018sound} and randomly sample 2 videos from MUSIC dataset to generate mixed audio for training and testing. %We use SGD optimizer with momentum 0.9 and learning rate 0.001 to train our network. For ResNet-18, we use learning rate 0.0001 since it is pre-trained on ImageNet. We train the network for 100 epochs and save the best model based on the validation error.

\begin{table}[h]
\small
   \begin{center}
   \vspace{-0.15in}
\caption{{\bf Cross dataset evaluation of audio separation.} 
  Evaluation of the model trained using the MUSIC dataset on the AudioSet-SingleSource dataset; using  ratio mask. 
  SAR is responsible for capturing absence of artifacts, therefore, it can be higher even when separation results are poor. }
  \label{table:audio_set}
  \vspace{0.03in}
  \begin{tabular}{l|ccc}
  \hline
    Methods & SDR ($\uparrow$) & SIR ($\uparrow$) & SAR ($\uparrow$)\\
    \hline
   Sound-of-Pixels~\cite{zhao2018sound} & 0.72 & 20.14 & \textbf{16.10}\\
   Ours(Mask coeff.) & 5.11 & 26.57 & 13.95 \\
   Ours(Mask coeff. + Seg. Net) & \textbf{7.19} & \textbf{29.98} & 12.15\\
    \hline
  \end{tabular}
  \end{center}
  \vspace{-0.35in}
\end{table}

\begin{table*}[h]
\small
   \begin{center}
   \vspace{-0.15in}
  \caption{{\bf Audio separation for unseen objects.} Toy experiment with cross dataset setting where the model never seen some instruments during training on MUSIC dataset. }
  \label{table:unseen_results}
  \vspace{0.03in}
  \begin{tabular}{l|ccc|ccc}\hline
     & \multicolumn{3}{c|}{Sound-of-Pixels~\cite{zhao2018sound}} & \multicolumn{3}{c}{Ours} \\
    \hline
    Instruments & SDR ($\uparrow$)  & SIR ($\uparrow$) & SAR ($\uparrow$) & SDR ($\uparrow$) & SIR ($\uparrow$) & SAR ($\uparrow$) \\
    \hline
    Banjo/Electric Guitar & 0.03 & 0.08 & {\bf 24.82} & {\bf 1.08} & {\bf 2.16} & 12.20 \\
    Saxophone/Marimba & 3.64 & 5.30 & 11.07 & {\bf 12.19} & {\bf 19.97} & {\bf 16.00} \\
    Cello/Electric Guitar & 0.79 & 0.81 & {\bf 28.20} & {\bf 2.07} & {\bf 3.57} & 9.33\\
    \hline
  \end{tabular}
  \end{center}
  \vspace{-0.25in}
\end{table*}

\begin{figure}[h]
  \centering
  \includegraphics[scale=0.36]{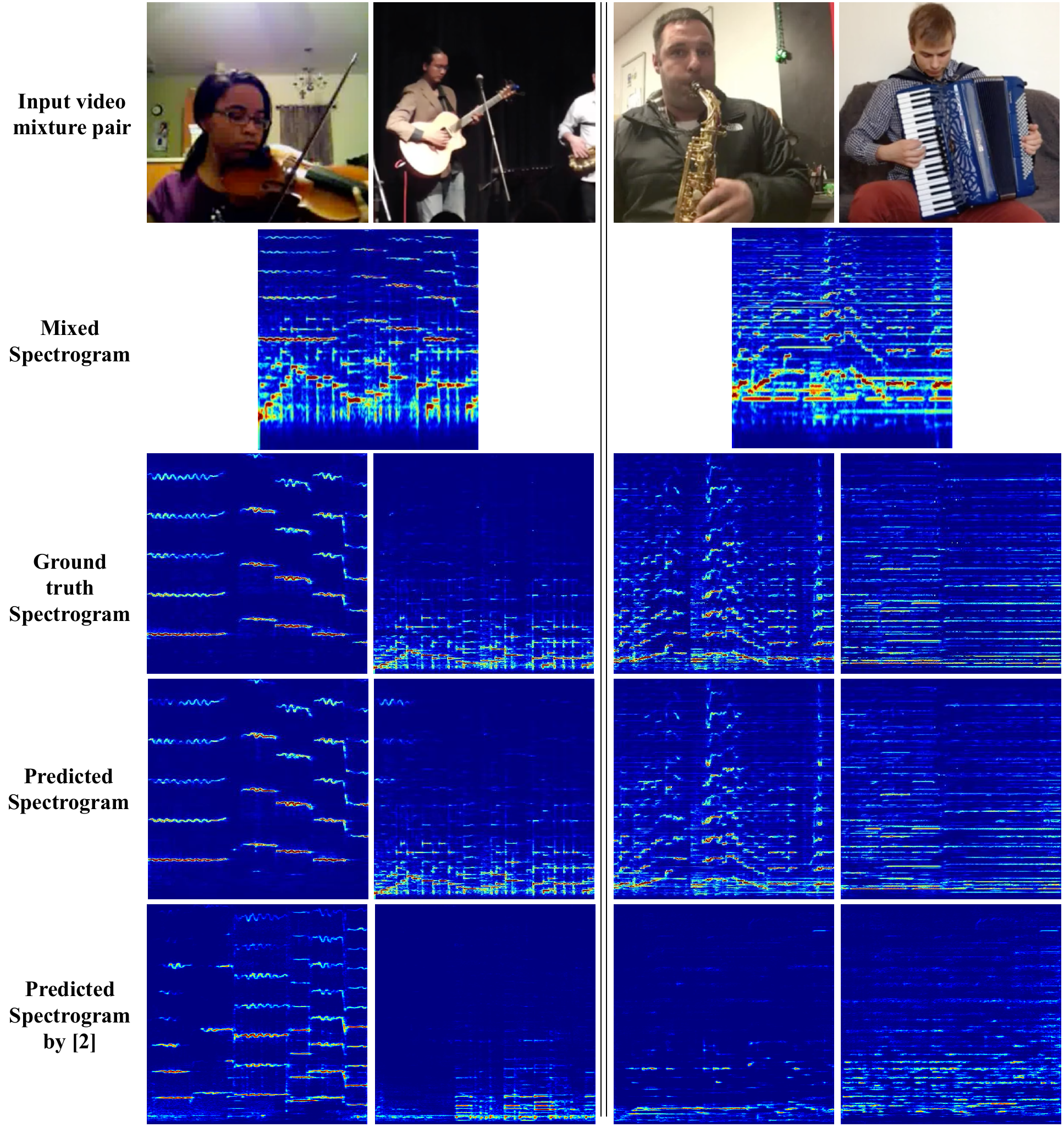}
  \vspace{-0.12in}  
  \caption{{\bf Qualitative audio separation results on MUSIC test set.}
  Test samples, our results and comparison with~\cite{gao2019co} are shown. See text for details and discussion.}
%   Qualitative results using our proposed sound source separation network guided by segmented visual object feature compared with~\cite{gao2019co}.}
  \label{fig:qualitative_results}
  \vspace{-0.20in}
\end{figure}

\vspace{-0.15in}
\subsection{Sound source separation and detection}
\vspace{-0.08in}
{\bf Evaluation Metrics.} To measure performance we use three widely used metrics for sound separation: Signal-to-Distortion Ratio (SDR), Signal-to-Interference Ratio (SIR), and Signal-to-Artifact Ratio (SAR). All the results are reported using widely used mir eval library~\cite{raffel2014mir_eval}. The baselines used to quantitatively compare our results (in Table~\ref{table:ex_results_final}) are described in supplementary material.
\vspace{0.03in}

\noindent
{\bf Visually guided sound source separation.} Table~\ref{table:ex_results_final} shows quantitative evaluation of experimental results on MUSIC dataset, using both binary and ratio masks. We also include sound separation results with and without weakly-supervised segmentation network, as an ablation, to show the importance of that module in our architecture. We note that additionally removing the mask-coefficient component effectively reduces our model to the Sound-of-Pixels~\cite{zhao2018sound} baseline -- the reason we do not include this variant. We note that improvements due to our decomposible construction of the mask are very significant ($7.26$ vs. $9.25$ in SDR using binary mask). The improvements due to weakly-supervised detection and segmentation is slightly more modest ($8.40$ vs. $9.14$ in SDR using ratio mask) but are still substantial. Consistent with \cite{gao2019co}, we find SDR and SIR metric to  be most informative.

Figure~\ref{fig:qualitative_results} shows corresponding qualitative results. The first and second rows illustrate randomly sampled video mixture pairs and corresponding spectrograms of the mixed sound. The third and fourth rows %show ground truth and predicted masks respectively; the fifth and sixth rows 
show ground truth and predicted separated spectrograms. Finally fifth row illustrated predicted spectrogram generated by running pre-train model from~\cite{gao2019co}\footnote{\url{https://github.com/rhgao/co-separation}}. One can clearly see that our method outperforms the state-of-the-art~\cite{gao2019co} in both quality and sharpness of resulting spectrograms. See supplementary material for additional ablations.

\begin{figure}[t]
  \centering
  \includegraphics[scale=0.33]{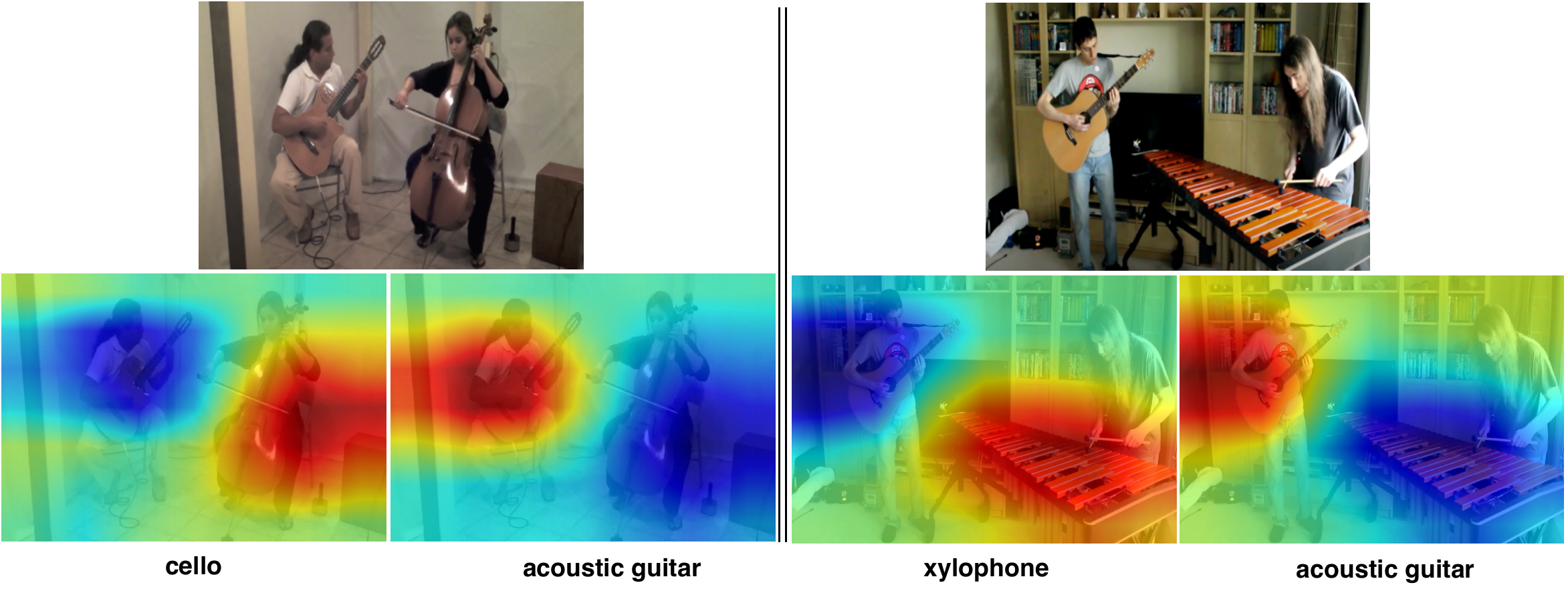}
  \vspace{-0.15in}
  \caption{{\bf Attended object map.} Attended maps (red higher attention) that correspond to object classes (in bottom). Result from our learned weakly-supervised segmentation network.}
  \label{fig:segmentation_results}
  \vspace{-0.15in}
\end{figure}

\vspace{0.03in}
\noindent
{\bf Sound object detection and segmentation.} 
Our object detection and segmentation utilizes a weakly-supervised network. Importantly, in addition to weakly-supervised loss, audio separation pathway, that depends on the resulting segmentations, provides additional regularization. We measure accuracy of our object detection network by computing multi-class classification accuracy on the MUSIC test set, as reported in Table~\ref{table:classification_results} as a function of the threshold $\tau$. Results illustrate that we can achieve high accuracy of up to $93.69$\% and that regularization with ratio mask variant of the audio network is consistently better for visual object detection. We visualize segmentation localization qualitatively (dataset does not contain spatial annotations for quantitative analysis) in Figure~\ref{fig:segmentation_results}. 

\vspace{0.03in}
\noindent
{\bf Cross-dataset experiments.} We also perform cross dataset testing to evaluate the generality of our method. 
We do so by measuring the performance of our proposed model, trained on MUSIC dataset, by applying it on the AudioSet-SingleSource dataset. The results are presented in Table~\ref{table:audio_set}. Note the nearly $10\times$ performance increase in SDR as compared to \cite{zhao2018sound}. 

\vspace{0.03in}
\noindent
{\bf Audio separation for unseen objects.} 
We also conduct a small experiment to see how the models perform for separating objects/instruments that the model has not seen during training. The results are presented in Table~\ref{table:unseen_results}. Here, the model never seen some instruments ({\em e.g.}, Banjo, Marimba) during training on MUSIC dataset but evaluated on those instruments from AudioSet-SingleSource dataset. In this case the model is relying on similarity of novel instruments to those used in training our model. 

\vspace{-0.17in}
\section{Conclusion}
\vspace{-0.10in}
In this paper, we introduce an end-to-end audio-visual co-segmentation network to separate and detect sound source without requiring additional supervision or bounding box proposal and solve the problem in a weakly supervised manner from large-scale unlabeled videos. Moreover, our mask coefficient generator facilitates separation conditioned on the output from the segmentation network. Both quantitative and qualitative results show the effectiveness of our proposed method compared to the existing state-of-the-art methods for sound source separation. %In future, we will incorporate motion feature in addition to visual feature to guide the sound source separation and detection task. 
\vspace{-0.10in}

\bibliographystyle{IEEEbib}
\bibliography{icme2021template_final}

\begin{thebibliography}{10}

\bibitem{cherry1953some}
E~Colin Cherry,
\newblock ``Some experiments on the recognition of speech, with one and with
  two ears,''
\newblock {\em JASA}, vol. 25, no. 5, pp. 975--979, 1953.

\bibitem{gao2019co}
Ruohan Gao and Kristen Grauman,
\newblock ``Co-separating sounds of visual objects,''
\newblock in {\em ICCV}, 2019, pp. 3879--3888.

\bibitem{zhao2019sound}
Hang Zhao, Chuang Gan, Wei-Chiu Ma, and Antonio Torralba,
\newblock ``The sound of motions,''
\newblock in {\em ICCV}, 2019, pp. 1735--1744.

\bibitem{zhao2018sound}
Hang Zhao, Chuang Gan, Andrew Rouditchenko, Carl Vondrick, Josh McDermott, and
  Antonio Torralba,
\newblock ``The sound of pixels,''
\newblock in {\em ECCV}, 2018, pp. 570--586.

\bibitem{owens2018audio}
Andrew Owens and Alexei~A Efros,
\newblock ``Audio-visual scene analysis with self-supervised multisensory
  features,''
\newblock in {\em ECCV}, 2018, pp. 631--648.

\bibitem{oktay2018attention}
Ozan Oktay, Jo~Schlemper, Loic~Le Folgoc, Matthew Lee, et~al.,
\newblock ``Attention u-net: Learning where to look for the pancreas,''
\newblock {\em arXiv}, 2018.

\bibitem{bolya2019yolact}
Daniel Bolya, Chong Zhou, Fanyi Xiao, and Yong~Jae Lee,
\newblock ``Yolact: real-time instance segmentation,''
\newblock in {\em ICCV}, 2019, pp. 9157--9166.

\bibitem{fitzgerald2016projet}
Derry Fitzgerald, Antoine Liutkus, and Roland Badeau,
\newblock ``Projet—spatial audio separation using projections,''
\newblock in {\em ICASSP}, 2016, pp. 36--40.

\bibitem{le2015deep}
Jonathan Le~Roux, John~R Hershey, and Felix Weninger,
\newblock ``Deep nmf for speech separation,''
\newblock in {\em ICASSP}, 2015, pp. 66--70.

\bibitem{owens2016ambient}
Andrew Owens, Jiajun Wu, Josh~H McDermott, William~T Freeman, and Antonio
  Torralba,
\newblock ``Ambient sound provides supervision for visual learning,''
\newblock in {\em ECCV}, 2016, pp. 801--816.

\bibitem{aytar2016soundnet}
Yusuf Aytar, Carl Vondrick, and Antonio Torralba,
\newblock ``Soundnet: Learning sound representations from unlabeled video,''
\newblock in {\em NIPS}, 2016, pp. 892--900.

\bibitem{xu2019recursive}
Xudong Xu, Bo~Dai, and Dahua Lin,
\newblock ``Recursive visual sound separation using minus-plus net,''
\newblock in {\em ICCV}, 2019, pp. 882--891.

\bibitem{pinheiro2015image}
Pedro~O Pinheiro and Ronan Collobert,
\newblock ``From image-level to pixel-level labeling with convolutional
  networks,''
\newblock in {\em CVPR}, 2015, pp. 1713--1721.

\bibitem{shen2017weakly}
Tong Shen, Guosheng Lin, Lingqiao Liu, Chunhua Shen, and Ian~D Reid,
\newblock ``Weakly supervised semantic segmentation based on
  co-segmentation.,''
\newblock in {\em BMVC}, 2017.

\bibitem{zhang2019decoupled}
Tianyi Zhang, Guosheng Lin, Jianfei Cai, Tong Shen, Chunhua Shen, and Alex~C
  Kot,
\newblock ``Decoupled spatial neural attention for weakly supervised semantic
  segmentation,''
\newblock {\em IEEE Transactions on Multimedia}, vol. 21, no. 11, pp.
  2930--2941, 2019.

\bibitem{gao20192}
Ruohan Gao and Kristen Grauman,
\newblock ``2.5 d visual sound,''
\newblock in {\em CVPR}, 2019, pp. 324--333.

\bibitem{griffin1984signal}
Daniel Griffin and Jae Lim,
\newblock ``Signal estimation from modified short-time fourier transform,''
\newblock {\em IEEE Transactions on Acoustics, Speech, and Signal Processing},
  vol. 32, no. 2, pp. 236--243, 1984.

\bibitem{he2016deep}
Kaiming He, Xiangyu Zhang, Shaoqing Ren, and Jian Sun,
\newblock ``Deep residual learning for image recognition,''
\newblock in {\em CVPR}, 2016, pp. 770--778.

\bibitem{spiertz2009source}
Martin Spiertz and Volker Gnann,
\newblock ``Source-filter based clustering for monaural blind source
  separation,''
\newblock in {\em DAFx}, 2009.

\bibitem{zhao2016loss}
Hang Zhao, Orazio Gallo, Iuri Frosio, and Jan Kautz,
\newblock ``Loss functions for image restoration with neural networks,''
\newblock {\em IEEE Transactions on Computational Imaging}, vol. 3, no. 1, pp.
  47--57, 2016.

\bibitem{gao2018learning}
Ruohan Gao, Rogerio Feris, and Kristen Grauman,
\newblock ``Learning to separate object sounds by watching unlabeled video,''
\newblock in {\em ECCV}, 2018, pp. 35--53.

\bibitem{raffel2014mir_eval}
Colin Raffel, Brian McFee, Eric~J Humphrey, Justin Salamon, Oriol Nieto, Dawen
  Liang, and Daniel~PW Ellis,
\newblock ``mir\_eval: A transparent implementation of common mir metrics,''
\newblock in {\em ISMIR}, 2014.

\end{thebibliography}

\end{document}